\documentclass[10pt,twocolumn,letterpaper]{article}
\usepackage{wacv}
\usepackage{wrapfig,lipsum,booktabs}
\usepackage{times}
\usepackage{epsfig}
\usepackage{graphicx}
\usepackage{amsmath}
\usepackage{amssymb}
\usepackage{subcaption}
\usepackage{boldline,multirow}
\usepackage{booktabs} 
\usepackage{comment}

\usepackage[pagebackref=true,breaklinks=true,letterpaper=true,colorlinks,bookmarks=false]{hyperref}
\wacvfinalcopy
\DeclareMathOperator*{\argmax}{argmax}

\ifwacvfinal\fi
\begin{document}
	
	\title{Combining Compositional Models and Deep Networks For \\Robust Object Classification under Occlusion}
	
	\author{
		Adam Kortylewski, 
		Qing Liu, 
		Huiyu Wang, 
		Zhishuai Zhang,
		Alan Yuille\\
		Johns Hopkins University\\
		{\tt\small \{akortyl1,qingliu,hwang157,zzhang99,ayuille1\}@jhu.edu}
	}
	\maketitle
	
\begin{abstract}
	Deep convolutional neural networks (DCNNs) are powerful models that yield impressive results at object classification. 
	However, recent work has shown that they do not generalize well to partially occluded objects and to mask attacks. 
	In contrast to DCNNs, compositional models are robust to partial occlusion,
	however, they are not as discriminative as deep models. 
	In this work, we combine DCNNs and compositional object models to retain the best of both approaches: a discriminative model that is robust to partial occlusion and mask attacks. 
	Our model is learned in two steps. First, a standard DCNN is trained for image classification. 
	Subsequently, we cluster the DCNN features into dictionaries.
	We show that the dictionary components resemble object part detectors and learn the spatial distribution of parts for each object class. 
	We propose mixtures of compositional models to account for large changes in the spatial activation patterns (e.g. due to changes in the 3D pose of an object). 
	At runtime, an image is first classified by the DCNN in a feedforward manner. 
	The prediction uncertainty is used to detect partially occluded objects, which in turn are classified by the compositional model.
	Our experimental results demonstrate that combining compositional models and DCNNs resolves a fundamental problem of current deep learning approaches to computer vision: 
	The combined model recognizes occluded objects, even when it has not been exposed to occluded objects during training, while at the same time maintaining high discriminative performance for non-occluded objects. 
\end{abstract}
	\begin{figure}
	\begin{subfigure}{0.49\linewidth}
		\centering
		\includegraphics[height=2.8cm]{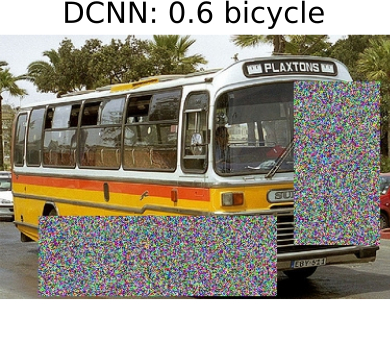}
		\caption{}
		\label{fig:intro-0}
	\end{subfigure}%
	\begin{subfigure}{0.49\linewidth}
		\centering
		\includegraphics[height=2.8cm]{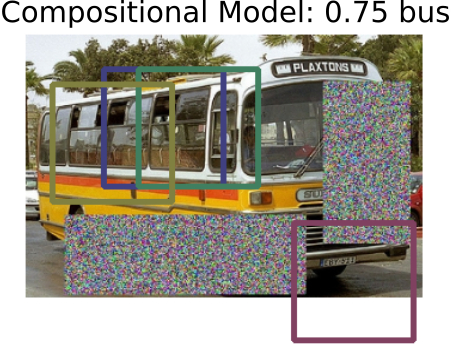}
		\caption{}
		\label{fig:intro-1}
	\end{subfigure}
	\\
	\begin{subfigure}{0.49\linewidth}
		\centering
		\includegraphics[height=2.8cm]{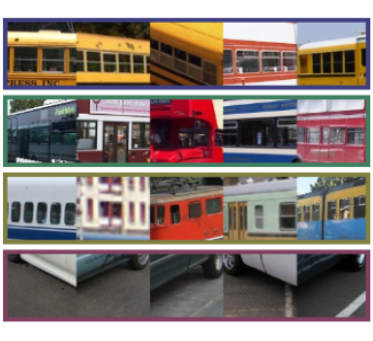}
		\caption{}
		\label{fig:intro-2}
	\end{subfigure}%
	\begin{subfigure}{0.49\linewidth}
		\centering
		\includegraphics[height=2.8cm]{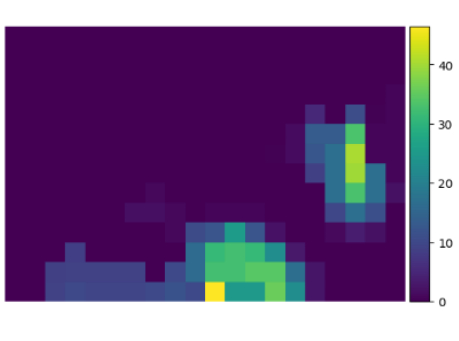}
		\caption{}
		\label{fig:intro-3}
	\end{subfigure}%
	\caption{Object classification under occlusion with DCNNs and compositional models. (a) The DCNN missclassifies the image as bicycle with low certainty. (b) The compositional model correctly classifies the image despite the strong partial occlusion. Intuitively, it can identify the object parts (colored rectangles in (b) and related parts from the training data in (c)) and ignore regions of the image which are inconsistent with the object model (d).} 
	\label{fig:intro}
\end{figure}	

\section{Introduction}
In natural images, objects are surrounded and partially occluded by other objects. 
Humans seem more robust to partial occlusion than current deep models \cite{hongru} (see our studies in Section \ref{sec:exp}).
One possible explanation is that it is unreasonable to assume that all possible occlusion patterns can be observed during training, because of their sheer number and variability.
Hence, a major difference between computer vision and other machine learning tasks is that in computer vision we cannot assume that the training and test data are sampled from the same underlying distribution. 
Thus, when deployed in the real-world, a vision system must generalize well beyond the training data. 
For example it should be able to recognize objects robustly in previously unseen illumination conditions (daylight vs dawn), poses (walking vs yoga) or partial occlusions. 
Prominent examples of vision systems failing to achieve this kind of generalization include fatal accidents caused by driver-assistance systems classifying a truck in an unusual pose as sky \cite{acc1} or failing to recognize a human that was partially occluded by a bicycle \cite{acc2}.
In this work, we address the task of classifying objects under partial occlusion. 
We propose a compositional model that can reason about partial occlusion, and hence is able to recognize partially occluded objects even when it has not been exposed to partial occlusion during training. Furthermore, we combine compositional models with a deep neural network into a model that is highly discriminative while also being robust to partial occlusion.

Deep convolutional neural networks (DCNNs) are powerful discriminative models that yield impressive results at object classification \cite{krizhevsky2012imagenet,simonyan2014very,he2016deep}. 
However, recent work has shown that DCNNs do not generalize well when objects are partially occluded \cite{wang2017detecting,hongru} and when they are exposed to mask attacks - adversarial examples where parts of the image are masked out \cite{fawzi2016measuring} (see also our experiments in Section \ref{sec:exp}). 
In contrast to deep models, compositional models have been shown to be robust to partial occlusion \cite{george2017generative,kortylewski2017model}, even if they have not seen partially occluded objects during training \cite{wang2017detecting,zhang2018deepvoting}. Compositional models explicitly represent an object in terms of parts and their spatial composition into a whole. 
The key benefit of such a compositional representation is two-fold: 
1) It makes possible to introduce an occlusion model that deactivates parts of the model, if they do not fit the data (i.e. if they are occluded by another object). 
2) The model can potentially explain its classification result in terms of where it has detected an objects' individual parts, as well as, where the object is occluded.
However, the major limitation of compositional models is that they lack the discriminative ability of deep learning approaches, because they are optimized for modeling the whole data distribution and not for discriminating between individual samples.
In this work, we propose to combine deep networks with compositional models, in order to get the best of both worlds, a highly discriminative model that is robust to partial occlusion and mask attacks. We make the following contributions in this paper:
\begin{itemize}
	\item \textbf{Learning compositional models from DCNN features.} In contrast to previous work which learns compositional models form the image pixels directly, we propose to learn them from DCNN features that are robust to nuisances such as illumination, background clutter and non-rigid deformations of parts. This enables us to represented complex objects in natural scenes, which is difficult to achieve with related approaches.
	\item \textbf{Generalization of compositional models to 3D objects.} We propose to model 3D objects with mixtures of compositional models, where each mixture component represents a particular viewpoint or 3D structure of an object. Our experiments show that mixture models are superior in terms of classification performance compared to single compositional models. 	
	\item \textbf{Combining compositional models and deep networks.} We propose to combine deep networks with compositional models into a model that retains high discriminative performance for non-occluded objects, while also being able to generalize well beyond what it has seen at training time in terms of partial occlusion. In our experiments, the proposed model outperforms a standard DCNN at classifying partially occluded objects by $13.9\%$ on the PASCAL3D+ dataset $19.4\%$ on MNIST digits and  in \textit{absolute} classification performance.
\end{itemize}

	\begin{figure*}
	\centering
	\includegraphics[height=4.5cm]{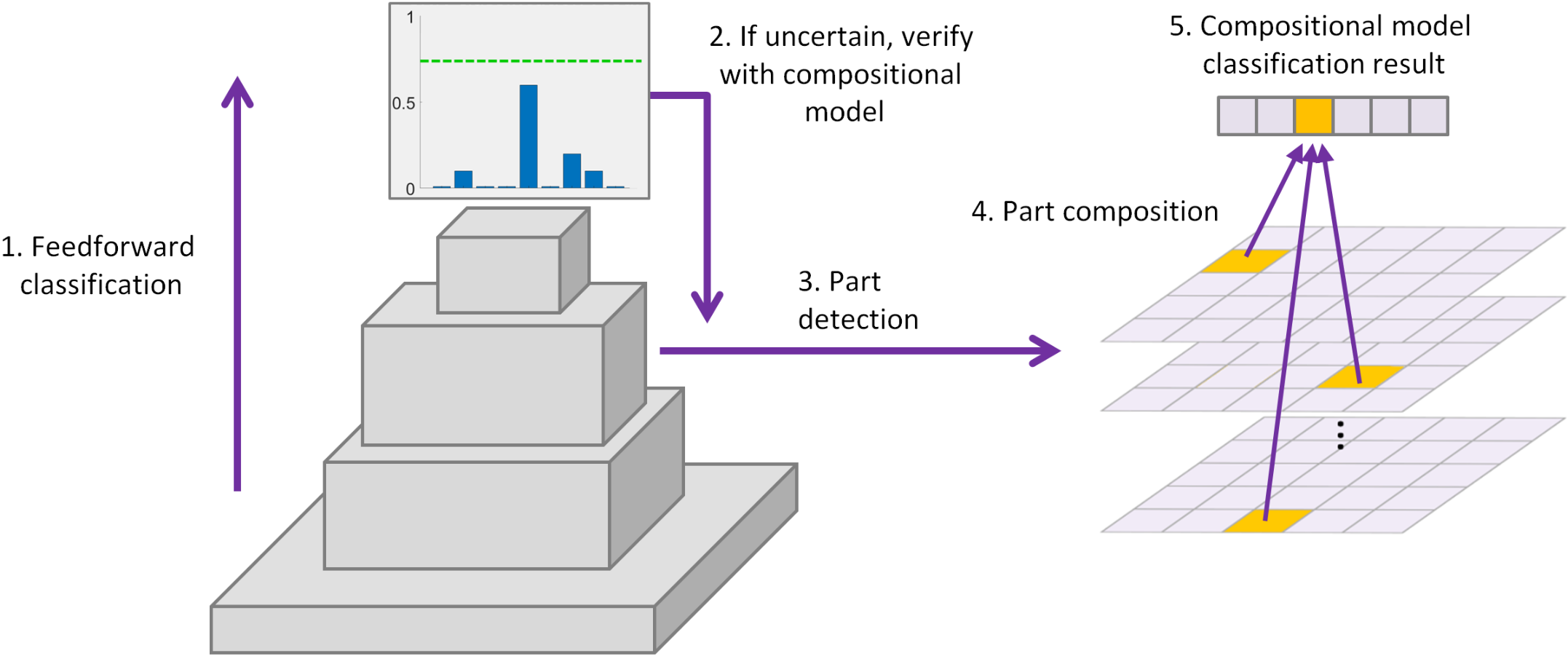}
	\caption{Overview of the proposed combination of DCNNs and compositional models.
		Our model has two branches, the DCNN branch (left) is highly discriminative but less robust, while the compositional model branch (right) is robust but less discriminative. 
		Both branches are integrated during inference. The model first classifies the input image with the DCNN-branch (1.). If the DCNN is uncertain about its prediction (2.), the test image is likely to be partially occluded. Hence, the initial prediction shall be verified with the compositional model. The parts of the compositional model are detected from the DCNNs feature map (3.) and combined (4.) into a robust prediction (5.).}
	\label{fig:model}
\end{figure*}

	\section{Related Work}
	\textbf{Classification under partial occlusion.} 
	In the context of deep learning, Fawzi and Frossard \cite{fawzi2016measuring} have shown that DCNNs are not robust to partial occlusion generated by masking out patches of the input image.
	In contrast to DCNNs, compositional models have been shown to be robust to partial occlusion. In particular they have been successfully applied for detecting partially occluded object parts \cite{wang2017detecting,zhang2018deepvoting} and for recognizing simple 2D shapes under partial occlusion \cite{george2017generative,kortylewski2016probabilistic,kortylewski2017model}.
	In this work, we propose a compositional model that can robustly classify 3D objects in natural scenes under strong partial occlusion.	 
	
	\textbf{Compositional object models.}
	Related works on compositional models for object classification \cite{jin2006context,zhu2008,fidler2014,dai2014unsupervised,kortylewski2017greedy} have proposed to learn the model parameters directly from image pixels. 
	The major challenge for these approaches is that their models need to explicitly account for nuisances such as illumination and object deformation in order to be robust to these nuisances.
	In this work, we propose to learn compositional models from the features of a DCNN. 
	DCNN features at higher layers of the network have been shown to be robust w.r.t. variation in the illumination, shape an appearance of an object \cite{zhou2014object,wang2015unsupervised,wang2017detecting}. 
	Hence, learning compositional model in terms of DCNN features instead of image pixels enables us to represent complex objects in natural scenes, without needing to model the underlying physical processes of the nuisances.
	
	\textbf{Combining compositional models and DCNNs.} 
	Liao et al. \cite{liao2016learning} propose to integrate the principles of compositionality into DCNNs by using a regularizer that encourages the feature representations of DCNNs to cluster during learning. They show that the resulting feature clusters resemble part detectors. Zhang et al. \cite{zhang2018interpretable} show that
	part detectors can be encouraged in DCNNs by restricting the activations in feature maps to have a localized distribution. 
	While these approaches have increased the explainability of the DCNN predictions, they have not been shown to enhance the robustness to partial occlusion. 
	Related approaches propose to regularize the convolution filters to be sparse \cite{tabernik2016towards}, or to enforce the activations in the feature maps to be disentangled for different objects \cite{stone2017teaching}. 
 	The key limitation of these approaches is that the compositional model is not explicit, but rather implicitly encoded within the parameters neural network. 
 	Thus, the resulting models remain black-box CNNs that are not robust to partial occlusion. 
 	In our proposed model the compositional model is explicit. Hence, it can be augmented with an occlusion model and become robust to partial occlusion, while also being able provide explanations of its' predictions in terms of where it perceives an objects parts and where it thinks the object is occluded.

	\section{A Robust Model Combining Deep Networks and Compositional Models}
In this section, we discuss how to combine compositional models and deep networks.
We present a dictionary-based compositional model including details of how the parameters of the model can be learned from data in Section \ref{sec:math}. 
In Section \ref{sec:occ}, we discuss how the compositional model can be made robust to partial occlusion. Finally, we discuss how a compositional model can be combined with a DCNN in Section \ref{sec:integration}.

\subsection{A Dictionary-Based Compositional Model of DCNN Features}
\label{sec:math}
Our long-term goal is to learn a generative model $p(F|y)$ of the DCNN features $F$ for an object class $y$, but we make simplifications (see next paragraph).
We define a feature map $F^l$ to be the output of a layer $l$ in a CNN. 
A feature vector $f^l_p \in \mathbb{R}^C$ is the vector of features in $F^l$ at position $p$, where $p$ is defined on the 2D lattice of the feature map and $C$ is the number of channels in the layer. 
Note that the spatial information from the image is preserved in the feature maps, thus a position $p$ on $F^l$ corresponds to a patch in the image. 
We omit the subscript $l$ in the remainder of this section because the layer from which the features are extracted is fixed in our model (e.g. $l=4$ for the layer $conv_4$).

\begin{figure}
	\centering
	\begin{subfigure}{0.24\linewidth}
		\centering
		\includegraphics[width=\linewidth]{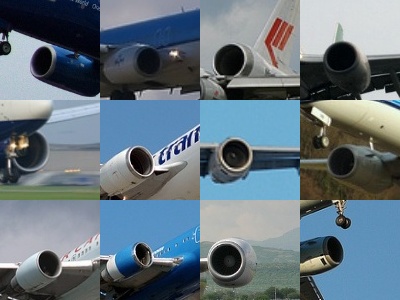}
		\caption{}
		\label{fig:parts-0}
	\end{subfigure}
	\begin{subfigure}{0.24\linewidth}
		\centering
		\includegraphics[width=\linewidth]{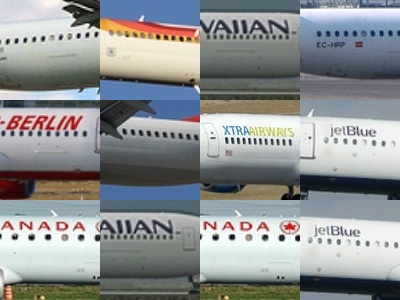}
		\caption{}
		\label{fig:parts-1}
	\end{subfigure}
	\begin{subfigure}{0.24\linewidth}
		\centering
		\includegraphics[width=\linewidth]{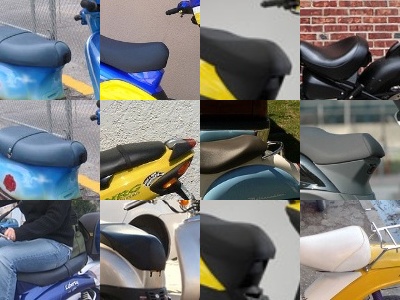}
		\caption{}
		\label{fig:parts-2}
	\end{subfigure}
	\begin{subfigure}{0.24\linewidth}
		\centering
		\includegraphics[width=\linewidth]{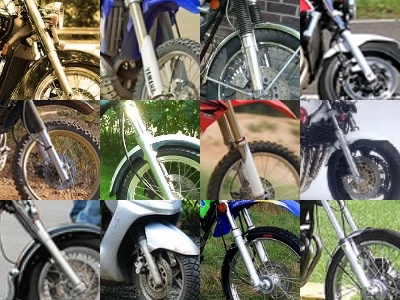}
		\caption{}
		\label{fig:parts-3}
	\end{subfigure}
	\caption{Illustration of part models by visualizing image patterns corresponding to the most likely feature vectors for a dictionary component. Note the variability in illumination, appearance and background suggesting robustness to these nuisances.}
	\label{fig:parts}
\end{figure}

\textbf{Learning dictionaries of DCNN features.} Modeling $p(F|y)$ is difficult because the feature maps are high dimensional and real valued. 
We propose to encode the feature maps with a dictionary $D=\{d_1,\dots,d_K\}$ that is learned by clustering the vectors from the feature maps of all training image $\{F^n | n=1,\dots,N\}$. 
We follow related work on learning dictionaries of DCNN features and use k-means for clustering \cite{wang2015unsupervised,wang2017detecting,liao2016learning}. 
In Figure \ref{fig:parts}, we illustrate some components $d_k$ of the learned dictionary $D$ by showing image patches that strongly activate this component. 
As previously observed in \cite{wang2015unsupervised,wang2017detecting}, the dictionary components activate image patches that are similar in appearance and often even share semantic meanings.
Note that the patches resemble image patterns that frequently re-occur for a particular class of images (e.g. Figure \ref{fig:parts-0} \& \ref{fig:parts-1} for the class airplane). 
Therefore, we refer to the components $d_k$ as \textit{parts}.

\textbf{Learning the spatial activation patterns of parts.} We encode the real valued feature vectors $f_p$ with a sparse binary vector $b_p$ by detecting the nearest neighbors of $f_p$ in the learned part dictionary $D$ using the cosine distance $g(\cdot|\cdot)$. 
Hence, the element $b_{p,k} = 1$ if $g(f_p,d_k)>\delta$.
Intuitively, $b_p$ encodes which parts of the dictionary $D$ are detected at position $p$ in the feature map $F$. 
Therefore, we refer to the resulting binary matrix $B$ as \textit{part detection map}.
We found that a threshold of $\delta=0.45$ causes $b_p$ to be sparse, while also at least one component is active at every position $p$ in $B$. 
We define a generative model of the part detection map as Bernoulli distribution:
\begin{equation}
\label{eq:fg}
p(B|\mathcal{A}_y) =  \prod_{p} p(b_p|\alpha_{p,y})= \prod_{p,k} \alpha_{p,k,y}^{b_{p,k}} (1-\alpha_{p,k,y})^{1-b_{p,k}}.
\end{equation}		
Where $\alpha_{p,k,y}$ is the probability that the part $d_k$ is active at position $p$ for the object class $y$, and thus $b_{p,k}=1$. 
Note that parts are assumed to be independently distributed which makes our model in spirit similar to bag of words models. However, the important difference is that the spatial position of the part detections are preserved in our model, hence capturing the spatial structure of the object.

\begin{figure*}[h]
	\centering
	\begin{subfigure}{0.25\linewidth}
		\centering
		\includegraphics[height=2.2cm]{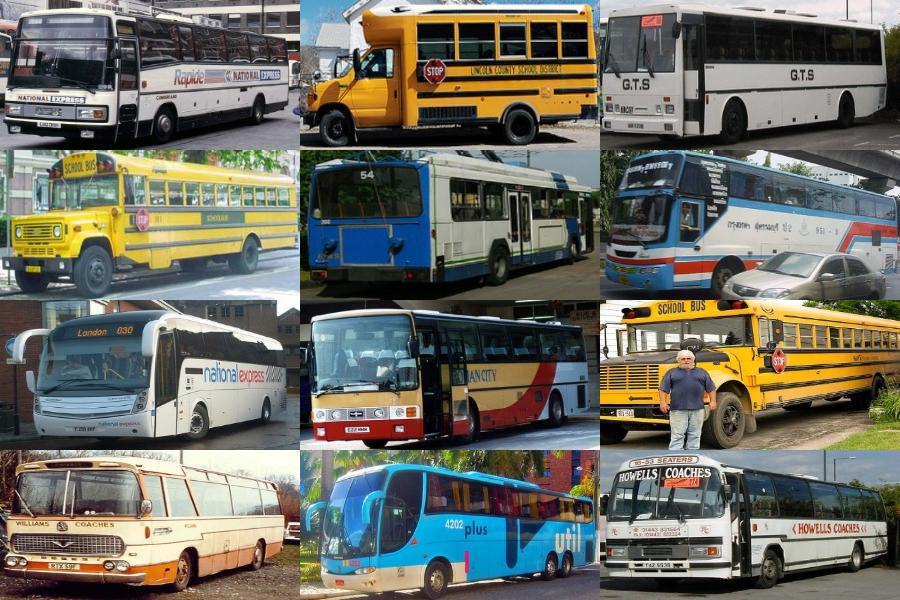}
		\caption{}
		\label{fig:clust2-0}
	\end{subfigure}%
	\begin{subfigure}{0.24\linewidth}
		\centering
		\includegraphics[height=2.2cm]{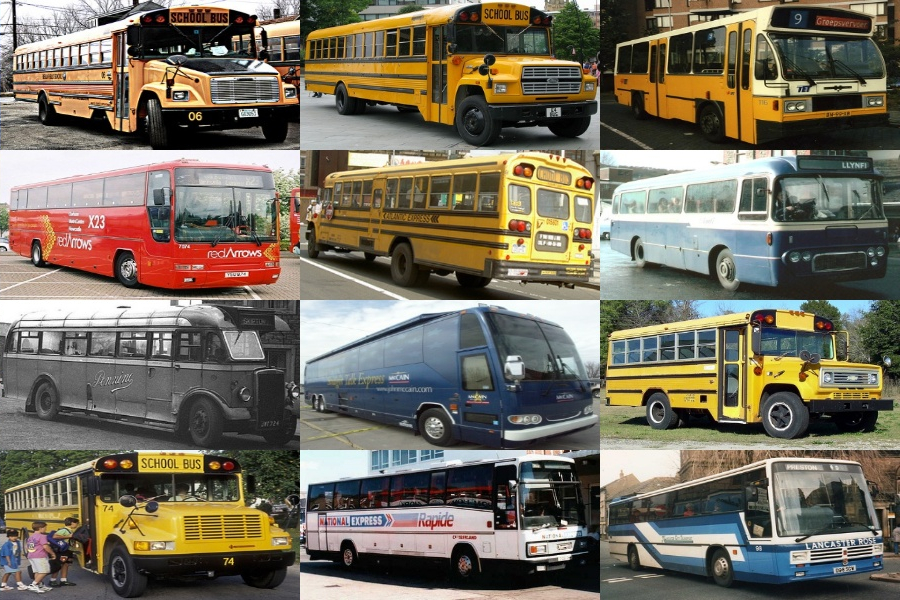}
		\caption{}
		\label{fig:clust2-1}
	\end{subfigure}
	\begin{subfigure}{0.24\linewidth}
		\centering
		\includegraphics[height=2.2cm]{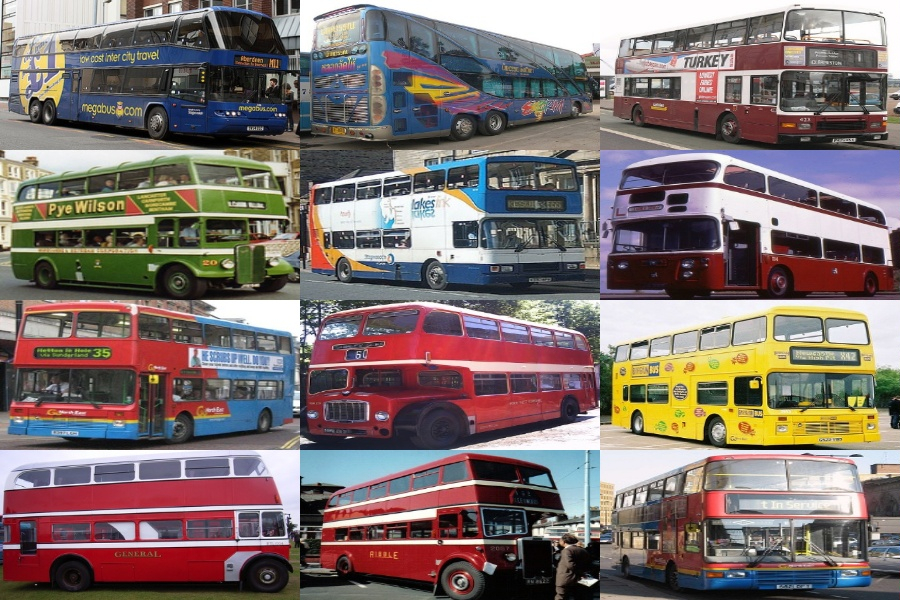}
		\caption{}
		\label{fig:clust2-2}
	\end{subfigure}	
	\begin{subfigure}{0.24\linewidth}
		\centering
		\includegraphics[height=2.2cm]{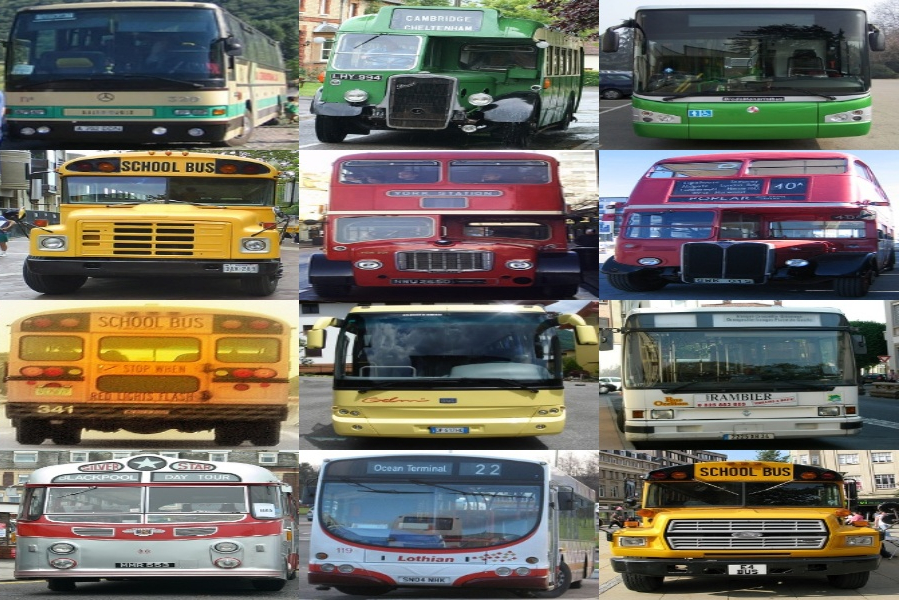}
		\caption{}
		\label{fig:clust2-3}
	\end{subfigure}
	\begin{subfigure}{0.24\linewidth}
		\centering
		\includegraphics[height=2.2cm]{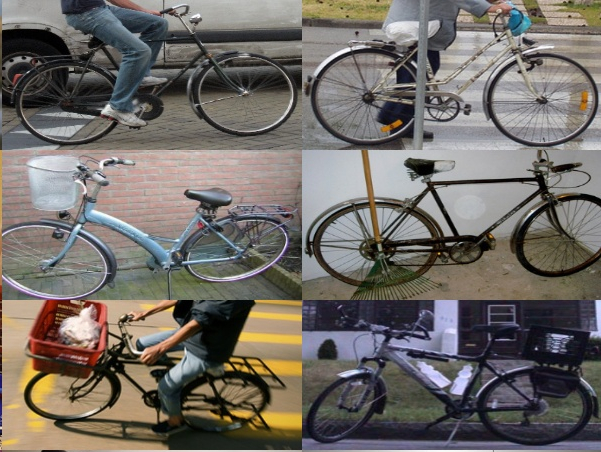}
		\caption{}
		\label{fig:clust-0}
	\end{subfigure}%
	\begin{subfigure}{0.24\linewidth}
		\centering
		\includegraphics[height=2.2cm]{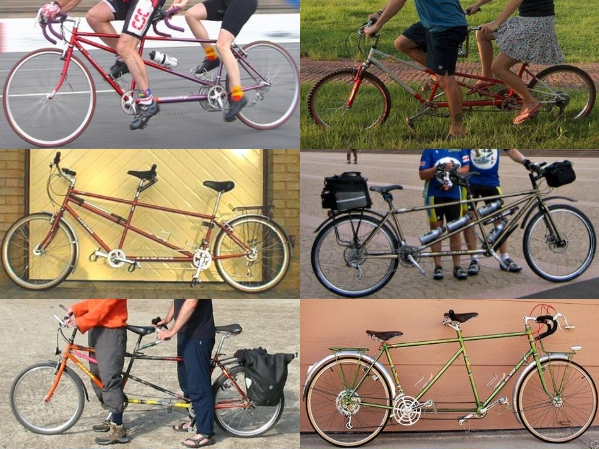}
		\caption{}
		\label{fig:clust-1}
	\end{subfigure}
	\begin{subfigure}{0.24\linewidth}
		\centering
		\includegraphics[height=2.2cm]{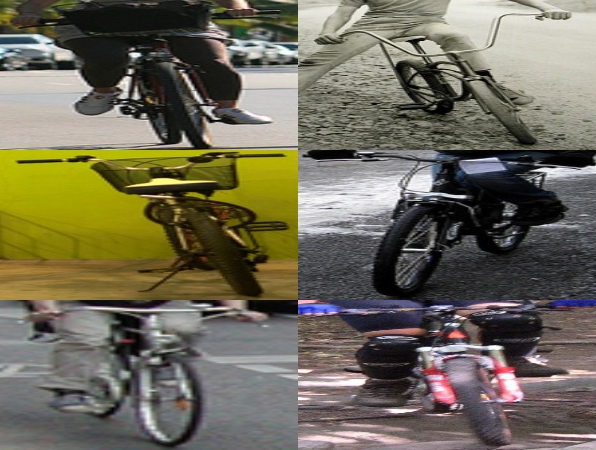}
		\caption{}
		\label{fig:clust-2}
	\end{subfigure}	
	\begin{subfigure}{0.24\linewidth}
		\centering
		\includegraphics[height=2.2cm]{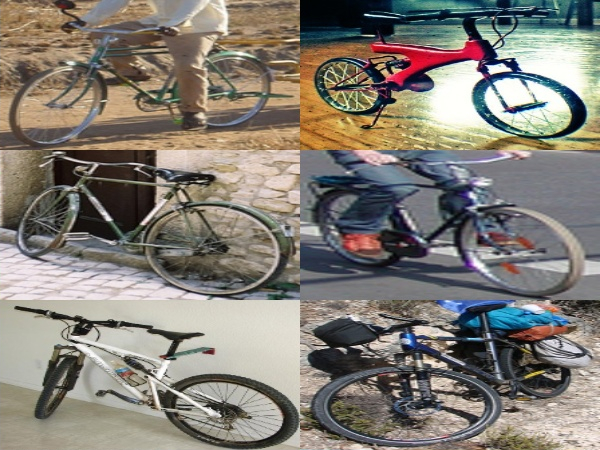}
		\caption{}
		\label{fig:clust-3}
	\end{subfigure}%
	\\
	\begin{subfigure}{0.24\linewidth}
		\centering
		\includegraphics[height=2.5cm]{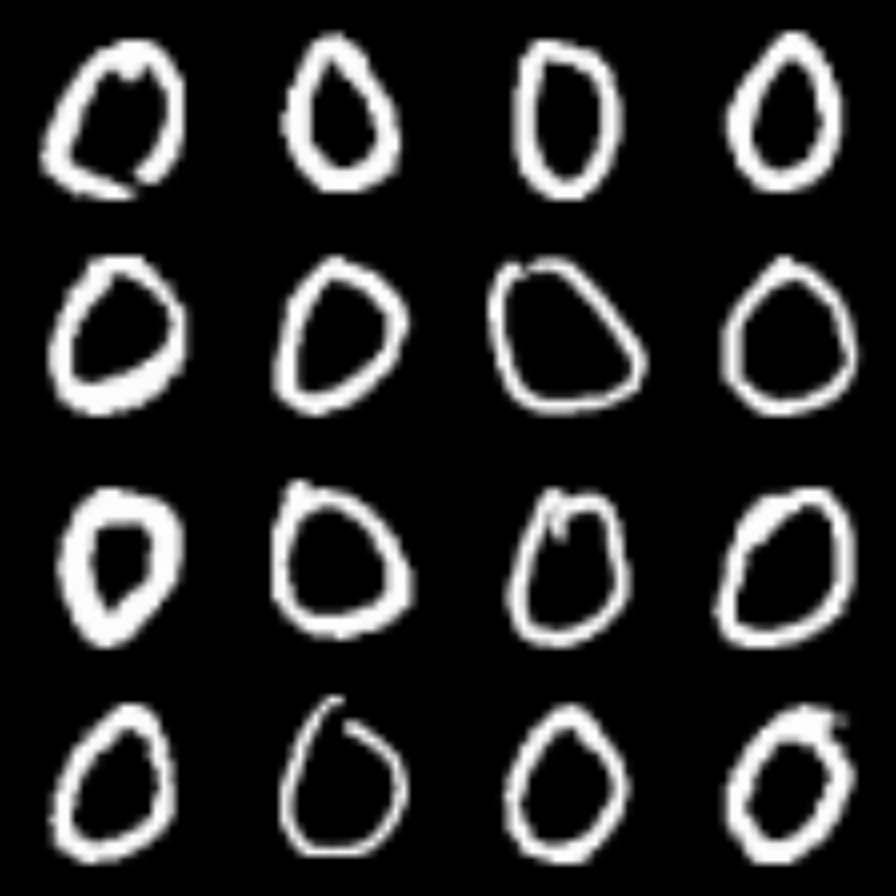}
		\caption{}
		\label{fig:clust2-0}
	\end{subfigure}%
	\begin{subfigure}{0.24\linewidth}
		\centering
		\includegraphics[height=2.5cm]{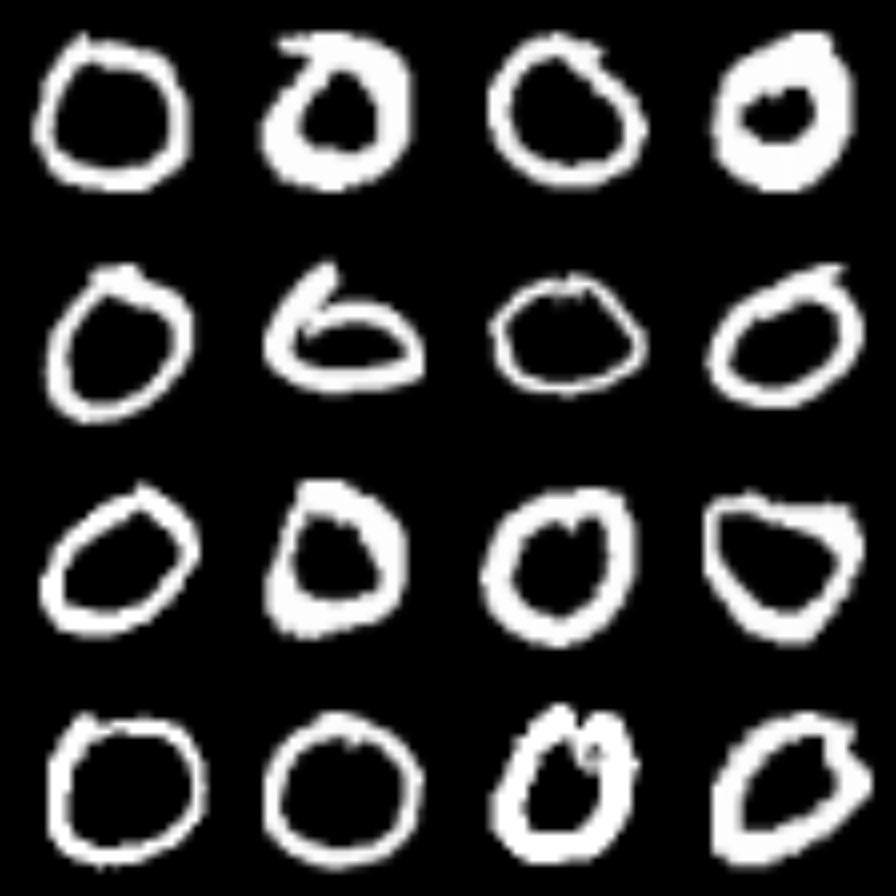}
		\caption{}
		\label{fig:clust2-1}
	\end{subfigure}
	\begin{subfigure}{0.24\linewidth}
		\centering
		\includegraphics[height=2.5cm]{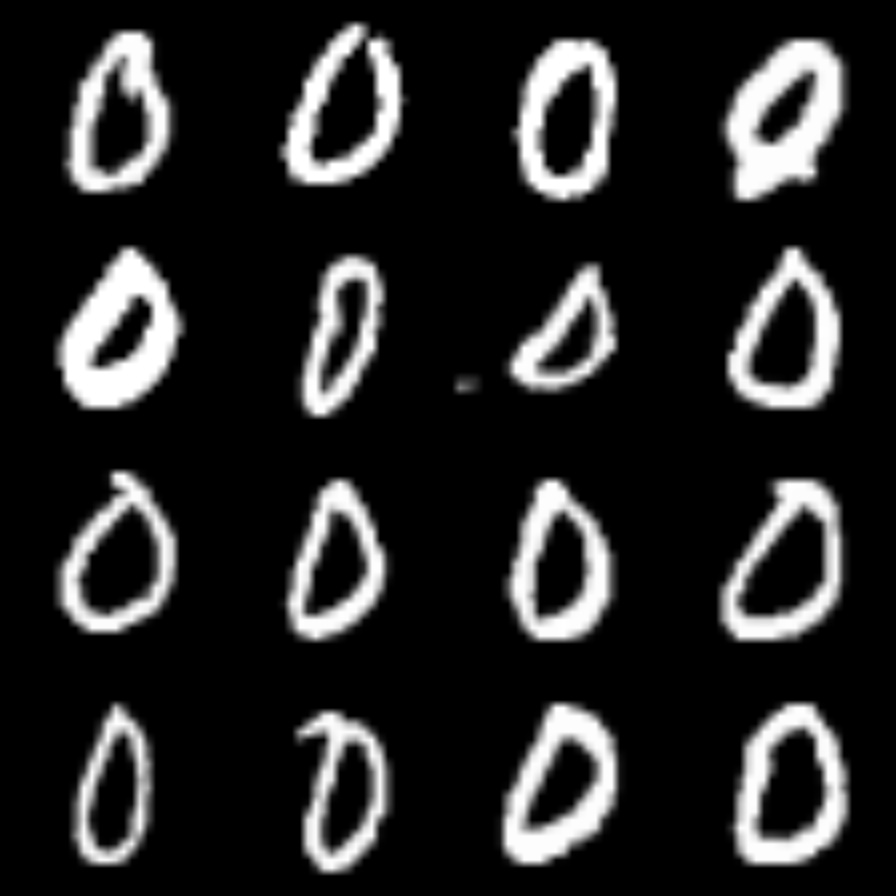}
		\caption{}
		\label{fig:clust2-2}
	\end{subfigure}	
	\begin{subfigure}{0.24\linewidth}
		\centering
		\includegraphics[height=2.5cm]{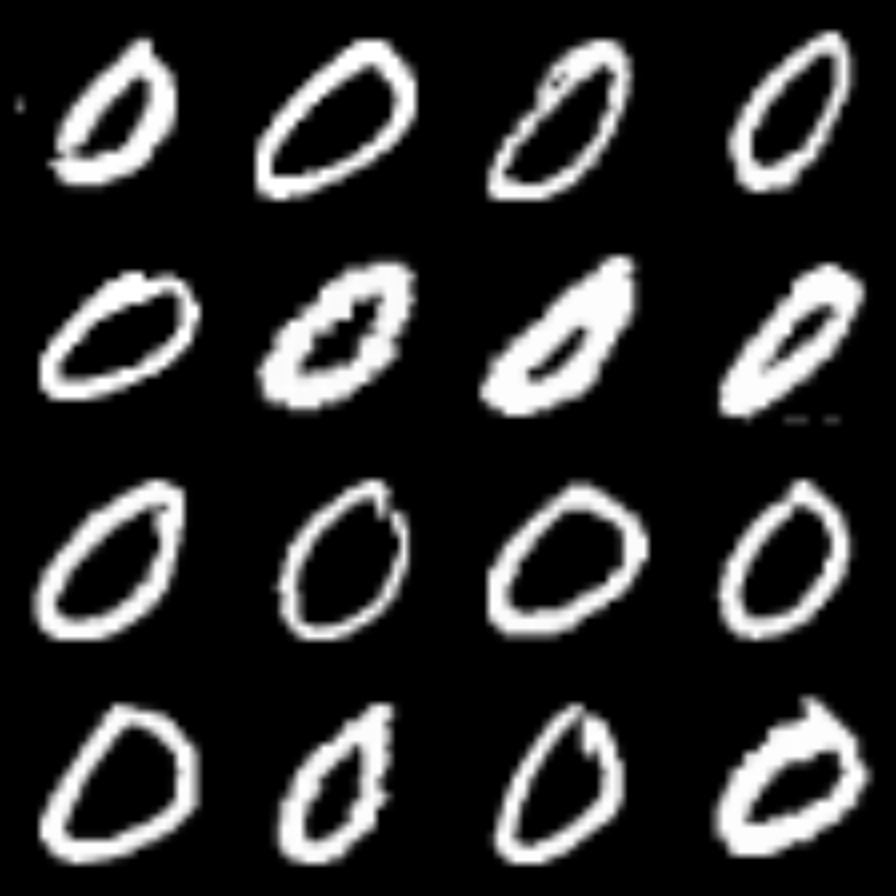}
		\caption{}
		\label{fig:clust2-3}
	\end{subfigure}		
	\caption{Visualization of mixture components obtained by performing clustering based on the part detection maps. Note how for busses and bicycles images with different 3D viewpoint or different structure (tandem) are approximately separated into different components (a-h), whereas MNIST images with different writing style are approximately separated.}
	\label{fig:clust}
\end{figure*}

\textbf{Mixture of compositional models.} Using the compositional model in Equation \ref{eq:fg} we can represent 2D objects (e.g. MNIST) as spatial composition of part detections. However, we are not able to represent 3D objects well (see results in Section \ref{sec:exp-ccn-occlusion}). 
The reason is that, due to independence assumption between parts in Equation \ref{eq:fg}, the model assumes that the spatial distribution of parts in $B$ is approximately the same.
This assumption does not hold for 3D objects, because e.g. by changing the 3D pose of an object the relative spatial distribution of parts changes strongly (e.g. the location of the tires of a car in the image change between the side view and a frontal view). 
In order to resolve this problem, we introduce mixtures of compositional models: 
\begin{equation}
p(B|\mathcal{A}_y;\mathcal{V}) = \prod_m p(B|\mathcal{A}^m_y)^{\nu_m}, \sum_m \nu_m = 1,\hspace{.1cm} \nu_m \in\{0,1\}.
\end{equation}
The intuition is that each mixture component $m$ will represent images of an object that have approximately the same spatial part distribution (i.e. similar viewpoint and 3D structure).
We learn the parameters of the Bernoulli distributions $A^m_y$ as well as the mixture assignment variables $\mathcal{V}$ using maximum likelihood estimation while alternating between estimating $A^m_y$ and $\mathcal{V}$. 
This approach essentially assumes that the variability of part detection maps within each a mixture component is smaller than between the mixture components.
To initialize the mixture assignments, we use spectral clustering with the hamming distance of the part detection maps of all training images $\{B^n|n=1,\dots,N\}$. 
The intuition is that objects with a similar viewpoint and 3D structure will have similar part activation patterns, and thus should be assigned to the same mixture component.
Figure \ref{fig:clust} illustrates the resulting cluster assignment after ten iterations with $m=4$ clusters for different objects. 
Note that objects with different viewpoints and spatial structure (e.g. tandems) are approximately separated into different clusters. 

\subsection{Augmenting the Compositional Model with an Occlusion Model}
\label{sec:occ}	
In natural images, objects are surrounded and partially occluded by other objects.
Partial occlusion of an object will change the part activation patterns in $B$ such that parts may be missing and other parts might be active at previously unseen location. 
The compositional model as described in Equation \ref{eq:fg} does not take this into account and thus will be distorted by partial occlusion (see experiments in Section \ref{sec:exp-ccn-occlusion}). 
However, modeling all of these “other objects” explicitly is computationally infeasible, because of their sheer number and variability. Hence, a common approach is to use an occlusion model \cite{kortylewski2017model}, where occluders are collectively modeled as locally independent clutter. 
The intuition behind an occlusion model is that at each position $p$ in the image either the object model $\mathcal{A}_y$ or a background model $\beta$ is active:
\begin{equation}
\label{eq:occ}
	p(B|\Theta_y) = \prod_{p} [p(b_p|\alpha_{p,y})p(z_p)]^{z_p}[p(b_p|\beta)(1-p(z_p))]^{1-z_p},
\end{equation}
where $\Theta_y = \{\mathcal{A}_y;\beta;\mathcal{Z}\}$. The binary variable $z_p \in \{0,1\}$ indicates if the object is visible at position $p$. The occlusion prior $p(z_p)$ could be learned or alternatively be set manually (see Section \ref{sec:exp}). The background model is defined as:
$	p(b|\beta) = \prod_{k} \beta_{k}^{b_{k}} (1-\beta_{k})^{1-b_{k}}$.
Here we assume that the background model is independent of the position in the image and thus it has no spatial structure. 
We estimate the background model as $\beta=\frac{1}{J}\sum_{j=1}^J b_j$ by sampling $J$ part detection vectors $b_j$ on a set of background images that do not contain one of the objects of interest. 
The maximum likelihood estimate of the occlusion variables $z_p$ can be computed efficiently due to the independence assumption in the occlusion model (Equation \ref{eq:occ}). 
Figure \ref{fig:exp} illustrates the positive values of the log-likelihood ratio between foreground and background model $\frac{p(b_p|\beta)}{p(b_p|\alpha_{p,y})}$. Note that the model can localize the occluder well. 
	
\subsection{Combining Compositional Models and DCNNs}
\label{sec:integration}

We combine the compositional model with the DCNN by first classifying an input image with both of their branches:
\begin{align}
y_{dcnn} &= \argmax_y p(y|I;W) ,\\
y_{cm}  &= \argmax_y p(B|\mathcal{A}_y;\beta;\mathcal{Z};\mathcal{V})	.
\end{align}
Our experiments show that the branches have complementary strengths and limitations. While the DCNN is highly discriminative for non-occluded objects, it performs poorly at classifying partially occluded objects, and vice-versa for the compositional model. Therefore, we combine both predictions into a final classification $y^*$ that retains the strengths of both branches, by setting $y^* = y_{dcnn}$ when $p(y_{dcnn}|I;W)>\tau$ and $y^* = y_{cm}$ else.
Here, $W$ are the parameters of the DCNN and $\tau$ is a threshold. The intuition is that if the DCNN is uncertain about its prediction (i.e. $p(y_{dcnn}|I;W)$ is low), then the input image is likely to be misclassified (e.g. due to occlusion) and hence should rather be classified by the compositional model.
Our experiments demonstrate that this approach successfully combines the complementary strengths of both branches.

	\section{Experiments}
\label{sec:exp}
\begin{figure}
	\centering	
	\begin{subfigure}{0.32\linewidth}
		\centering
		\includegraphics[height=2.4cm]{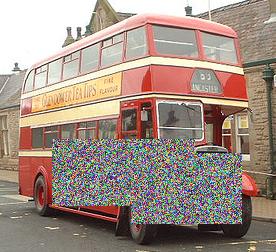}
		\caption{}
		\label{}
	\end{subfigure}
	\begin{subfigure}{0.32\linewidth}
		\centering
		\includegraphics[height=2.4cm]{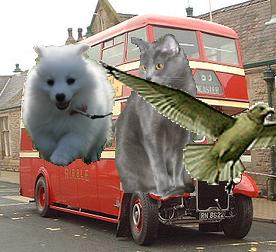}
		\caption{}
		\label{fig:occarea-obj}
	\end{subfigure}
	\begin{subfigure}{0.32\linewidth}
		\centering
		\includegraphics[height=2.4cm]{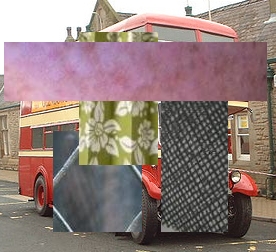}
		\caption{}
		\label{}
	\end{subfigure}
	\\
	\begin{subfigure}{0.32\linewidth}
		\centering
		\includegraphics[height=2.4cm]{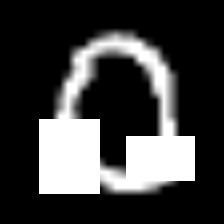}
		\caption{}
		\label{}
	\end{subfigure}
	\begin{subfigure}{0.32\linewidth}
		\centering
		\includegraphics[height=2.4cm]{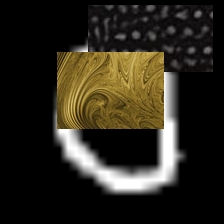}
		\caption{}
		\label{}
	\end{subfigure}
	\begin{subfigure}{0.32\linewidth}
		\centering
		\includegraphics[height=2.4cm]{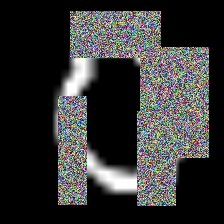}
		\caption{}
		\label{}
	\end{subfigure}\\
	\caption{Visualization of synthetic partial occlusions for natural objects (a-c) and MNIST digits (d-f) with varying amount of occlusion area: (a\&d) 20-40\% occlusion. (b\&e) 40-60\% occlusion. (d\&e) 60-80\% occlusion. We simulate different types of occlusion appearances: random noise (a\&f), natural objects (b), textures (c\&e) or white color (d).}	
	\label{fig:occarea}
\end{figure}

We evaluate our model at the task of object classification on partially occluded MNIST digits \cite{lecun1998mnist} and vehicles from the PASCAL3D+ dataset \cite{xiang2014beyond}. 
We simulate partial occlusion (Figure \ref{fig:occarea}) by masking out patches in the images and filling them with random noise, textures, or constant white color . 
For the PASCAL3D+ vehicles we additionally use the images provided in the VehicleSemanticPart dataset \cite{wang2015unsupervised}, where partial occlusion was simulated by superimposing segmented objects over the target object (Figure \ref{fig:occarea-obj}). 
Note that the objects used to simulate partial occlusion are different from the objects that the model has to discriminate.
We define different occlusion levels which correspond to increasing amounts of occlusion based on the object segmentation masks provided in the PASCAL3D+ dataset as well as threshold segmentations of the MNIST digits. We quantify how recognizable the occluded objects are by reporting the average performance of five subjects that were asked to perform every type of experiment in Table \ref{tab:occ} (total of $920$ human classifications). 

\textbf{Training details and parameter settings.} 
We train and evaluate our models on the standard train/test splits as defined in the respective datasets. 
For the PASCAL3D+ data we follow the setup as proposed in \cite{wang2015unsupervised}. Thus, the task is to discriminate between 12 objects during training, while at test time the six vehicle categories are tested.
If not differently stated, the models are trained on non-occluded objects, while at test time they are exposed to objects with different levels of partial occlusion.
The DCNN has a VGG-16 architecture \cite{simonyan2014very} and was pre-trained for object classification on the ImageNet dataset \cite{deng2009imagenet}. 
For training the compositional model, all images are resized such that their short edge has a size of $224$ pixels. We extract the features form the $pool4$ layer of the DCNN. 
The mixture models have $m=4$ components. We learn $50$ dictionary components for each object class, thus the dictionary $D$ has $K=500$ for the MNIST dataset and $K=600$ components for the PASCAL3D+ dataset. 
We learn a background model $\beta$ for each of the four types of occluders and use a threshold of $\tau = 0.6$ for the combination of the two branches. For experiments including an occlusion model, we use a prior of $p(z)=0.7$ that is the same for all positions $p$.

\begin{table*}
	\small
	\centering
	\tabcolsep=0.11cm
	\begin{tabular}{lV{2.5}cV{2.5}c|c|c|cV{2.5}c|c|c|cV{2.5}c|c|c|cV{2.5}c}
		\multicolumn{15}{c}{\textbf{PASCAL3D+ Classification under Occlusion}} \\
		\hline
		Occ. Area 					         & \textbf{0\%} & \multicolumn{4}{cV{2.5}}{\textbf{Level-1: 20-40\%}} & \multicolumn{4}{cV{2.5}}{\textbf{Level-2: 40-60\%}} & \multicolumn{4}{cV{2.5}}{\textbf{Level-3: 60-80\%}} & Mean \\
		\hline
		Occ. Type 	& - & w & n & t & o & w & n & t& o & w & n & t & o &-\\    		
		\hline  
		VGG 		& \textbf{98.6} & \textbf{96.8} & 94.9 & 96.0 & 87.9 &89.2&84.2&86.2&66.3&50.2&43.8&45.8&42.7&75.6\\
		\hline
		Comp &94.6&86.7&93.0&93.0&93.3&59.0&85.5&86.5&87.6&22.5&60.7&63.8&78.1&77.3\\
		\hline
		CompOcc& 89.4&90.4&89.2&88.6&89.2&85.3&86.6&84.8&87.8&70.0&77.4&72.0&84.2&84.2\\
		\hline  
		CompMix &93.6&80.6&90.7&89.8&92.0&58.8&83.0&83.0&88.5&26.1&59.6&65.1&84.3&76.6	\\
		\hline			
		CompMixOcc		&92.1&92.7&92.3&91.7&92.3&87.4&89.5&88.7&90.6&70.2&80.3&76.9&87.1&87.1\\	\hline		
		CompOccMix+VGG	&98.3&\textbf{96.8}&\textbf{95.9}&\textbf{96.2}&\textbf{94.4}&\textbf{91.2}&\textbf{91.8}&\textbf{91.3}&\textbf{91.4}&\textbf{71.6}&\textbf{80.7}&\textbf{77.3}&\textbf{87.2}&\textbf{89.5}\\
		\hline
		\hline		
		Human & 100.0& \multicolumn{4}{cV{2.5}}{100.0}& \multicolumn{4}{cV{2.5}}{100.0}  & \multicolumn{4}{cV{2.5}}{98.3}& 99.5
		\vspace{.2cm}
	\end{tabular}
	\begin{tabular}{lV{2.5}cV{2.5}c|c|cV{2.5}c|c|cV{2.5}c|c|cV{2.5}c}
		\multicolumn{12}{c}{\textbf{MNIST Classification under Occlusion}} \\
		\hline
		Occ. Area 					& \textbf{0\%}& \multicolumn{3}{cV{2.5}}{\textbf{Level-1: 20-40\%}}& \multicolumn{3}{cV{2.5}}{\textbf{Level-2: 40-60\%}}  & \multicolumn{3}{cV{2.5}}{\textbf{Level-3: 60-80\%}}& Mean \\
		\hline
		Occ. Type& - & w & n & t & w & n & t& w & n & t &-\\    		
		\hline  
		VGG 		&\textbf{99.5}&78.5&63.0&69.3&54.2&39.4&41.7&23.5&17.5&17.3&50.4\\
		\hline			
		CompOcc 		&89.7&77.7&76.9&77.8&67.6&66.2&67.6&42.5&40.6&42.5&64.9\\
		\hline			
		CompMixOcc 		&92.9&82.4&81.4&82.1&71.8&70.9&72.5&43.2&40.8&44.0&68.2\\		
		\hline			
		CompOccMix+VGG &99.1&\textbf{85.2}&\textbf{82.3}&\textbf{83.4}&\textbf{72.4}&\textbf{71.0}&\textbf{72.8}&\textbf{43.5}&\textbf{41.2}&43.0&\textbf{69.4}\\
		\hline
		\hline				
		Human & 100.0& \multicolumn{3}{cV{2.5}}{92.7}& \multicolumn{3}{cV{2.5}}{91.3}  & \multicolumn{3}{cV{2.5}}{64.0}& 84.4
	\end{tabular}
	\caption{Classification results for PASCAL3D+ and MNIST with different levels of occlusion (0\%,20-40\%,40-60\%,60-80\% of the object are occluded), different types of occlusion (w=white boxes, n=noise boxes, t=textured boxes, o=natural objects) and human classification baselines.}	
	\label{tab:occ}
	\begin{tabular}{lV{2.5}c|c|cV{2.5}c|c|cV{2.5}c}
		\toprule
		\multicolumn{8}{c}{\textbf{Training with Occlusion Bias on MNIST 20-40\%}} \\
		\hline
		Occ. Bias	& \multicolumn{3}{cV{2.5}}{\textbf{Left-Half}}& \multicolumn{3}{cV{2.5}}{\textbf{Right-Half}}  & Mean \\
		\hline
		Occ. Type  & w & n & t & w & n & t& -\\    		
		\hline  
		VGG\_R  &76.2&71.7&73.6&97.5&97.4&97.3&85.4\\
		\hline		 	
		CompOccMix+VGG\_R &83.3&82.0&83.0&97.3&97.1&96.9&90.0\\		
		\hline  
		\hline  
		VGG\_R\_W  			   &80.8&63.9&67.4&97.1&93.4&93.8&82.7\\		
		\hline		 	
		CompOccMix+VGG\_R\_W &86.4&82.5&82.4&96.9&93.4&94.1&89.3\\				
	\end{tabular}	
	\caption{Classification results when the occluders in the training images are biased to occur only in the right half of the image (*\_R) and when additionally they are biased to have white color (*\_R\_W).}
	\label{tab:bias}
\end{table*}	

\begin{figure*}
	\begin{subfigure}{\linewidth}
	\centering			
	\includegraphics[height=2.55cm]{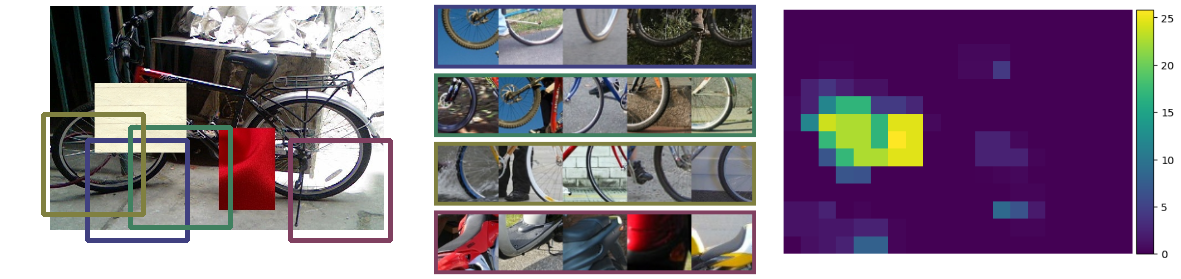}
		\caption{}
		\label{fig:exp-0}
	\end{subfigure}
	\begin{subfigure}{\linewidth}
	\centering		
	\includegraphics[height=2.55cm]{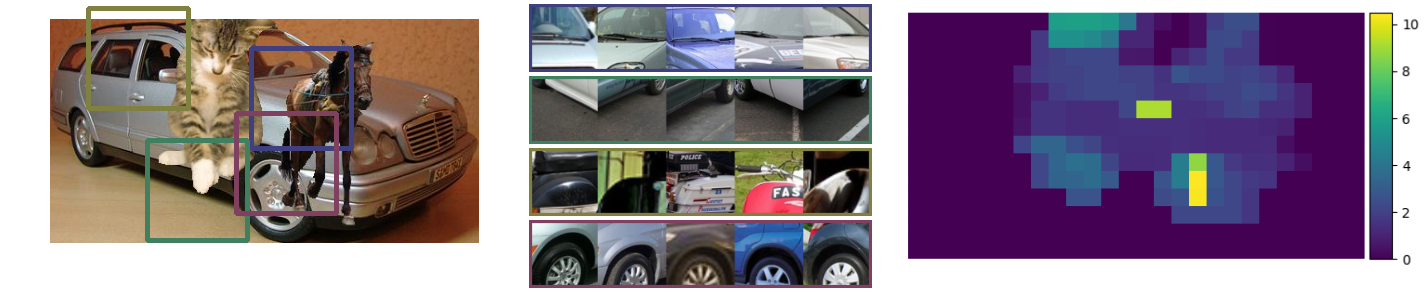}
	\caption{}
	\label{fig:exp-2}
\end{subfigure}
	\begin{subfigure}{\linewidth}
	\centering		
	\includegraphics[height=2.55cm]{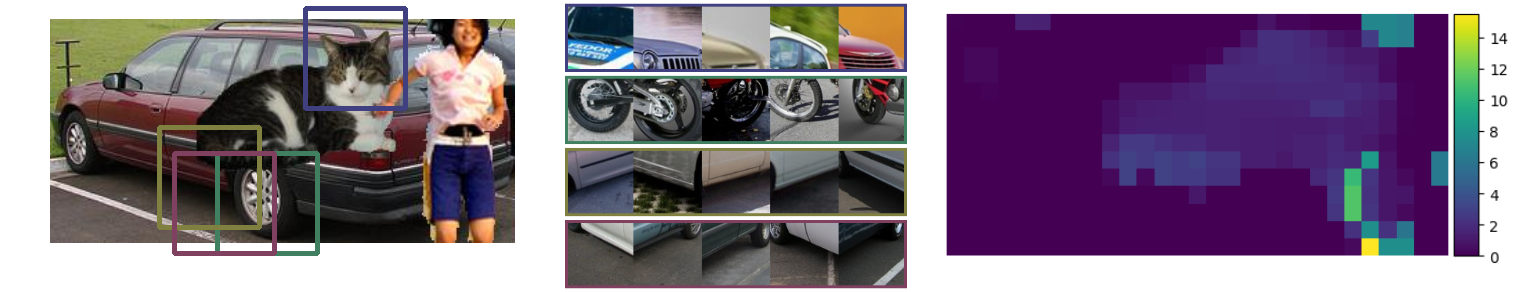}
	\caption{}
	\label{fig:exp-3}
\end{subfigure}
\begin{subfigure}{\linewidth}
	\centering		
	\includegraphics[height=2.55cm]{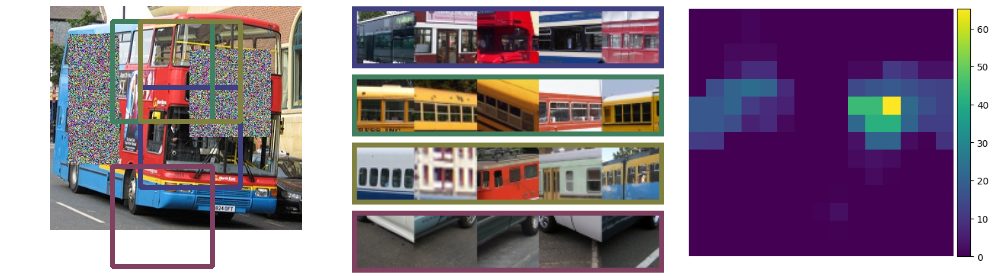}
	\caption{}
	\label{fig:exp-4}
\end{subfigure}
\caption{Illustration how a compositional model can provide explanations of its' prediction in terms of where it perceives the object parts (colored rectangles in the left image and related parts from the training data in the middle image) and where it thinks the object is occluded (right). To generate the occlusion map, we plot the positive log-likelihood ratio between background model and the compositional model.}
\label{fig:exp}
\end{figure*}

\subsection{DCNNs Do Not Generalize Well Under Partial Occlusion}
\label{sec:exp-dcnn-occlusion}
The classification results in Table \ref{tab:occ} show that the VGG network does not generalize well under partial occlusion, when it was not exposed to partially occluded objects during training. For the PASCAL3D+ data, the DCNN achieves a good performance for non-occluded objects and level-1 mask attacks. While for stronger levels of occlusion the performance drops by more than $10\%$. Note that for natural occluders the performance decrease is much higher at level-1 and level-2 compared to mask attacks.

In large-scale datasets, we can expect that some amount of partial occlusion will be present in the data. 
However, it is well known that the variability in large datasets is often biased. 
Thus, the location of the partial occlusions might also be affected by dataset bias. 
We simulate this by training the DCNN with MNIST images with a combination of non-occluded images and images where the occluders occur only in the right half of the image at training time (VGG\_R), while at test time they can occur all over the image. 
The classification results in Table \ref{tab:bias} show that the DCNN can classify partially occluded objects well, when the partial occlusion occurs at locations it has observed during training (Right-Half). 
However, it cannot generalize well when the object is occluded at previously unseen spatial positions (Left-Half). 
We simulate an even more severe bias by restricting the occluders to also have a biased appearance (white masks only) in addition to having a biased location (VGG\_R\_W). We observe that the performance drops for previously unseen appearances (noise and textures) at all locations in the image, while it increases for the occluders with the same appearance at previously unseen positions (white masks in the left half).
Hence, we observe a complex relation between biases in the training data and the classification performance that demands further studies.

Overall, our experiments show that DCNNs do not generalize well to previously unseen partial occlusion. 
However, it is important for computer vision systems to generalize away from the training data in terms of partial occlusion, because in real-world applications computer vision systems are almost always exposed to dataset bias in terms of partial occlusions.

\subsection{The Proposed Model Classifies Partially Occluded Objects Robustly}
\label{sec:exp-ccn-occlusion}

\textbf{PASCAL3D+.} The results in Table \ref{tab:occ} show that our proposed combination of compositional models and DCNNs outperforms the VGG network at classifying partially occluded objects for all levels and all types of occlusion, while retaining comparable performance for non-occluded objects. For level-1 mask attacks the performance of VGG and our combined model (CompOccMix+VGG) is comparable, while it becomes more prominent for level-2 and level-3 attacks with a mean \textit{absolute} performance gain of $4.9\%$ and $29.9\%$ respectively. The absolute performance gain is even more prominent if the occluders are real objects (level-1: $6.5\%$; level-2: $25.1\%$; level-3:$44.5\%$). Note that while our proposed model has not been exposed to partial occlusion at training time it is still able to classify partially occluded objects with exceptional accuracy.

\textbf{MNIST.} For the MNIST data we can observe similar generalization patterns as we have observed for PASCAL3D+. 
Our model is able to classify the partially occluded digits better than the VGG network, with a mean absolute performance gain of $12.1\%$ for level-1, $25.7\%$ for level-2 and $27.5\%$ for level-3 occlusions. 
Additionally, when the occlusions during training have a bias in the spatial positions and/or the appearance, our model generalizes much better to previously unseen partial occlusions than the VGG network (Table \ref{tab:bias}).
Interestingly, the mixture of compositional models (CompOccMix) also provides a performance increase for the two dimensional MNIST digits compared to a single compositional model (CompOcc). 
In Figure \ref{fig:clust}, we show that each mixture focuses on a particular writing style of a digit, suggesting that it can better approximate the distribution of handwritten digits and hence is able to better discriminate between them.

In summary, we observe that a combination of compositional models and DCNNs generalizes much better to previously unseen data in terms of partial occlusion compared to using a standard DCNN only, while having comparable performance on data that is similarly distributed as the one observed during training.

\textbf{Ablation study.}
Table \ref{tab:occ} contains a series of ablation experiments on the PASCAL3D+ data. 
On average, single compositional models (Comp) as well as mixtures of compositional models (CompMix) perform as good as a DCNN. 
While they perform worse for images without occlusion and for level-1 occlusions, they are better for level-2 and level-3 occlusions compared to the DCNN. 
Hence, we can clearly observe the complementary strength and weakness of both types of models. 
When augmented with on occlusion model (CompOcc and CompMixOcc) the compositional models clearly outperform VGG in absolute performance by $8.6\%$ and $11.5\%$ respectively. 
Note that the mixture of compositional models performs superior compared to a single compositional model when they are augmented with an occlusion model. 
The combination of the VGG branch and the occlusion-aware mixture (CompOccMix+VGG) improves the performance for all experiments on partially occluded objects, while retaining comparable performance to the VGG model for non-occluded objects. 
Note the mutual benefit of integrating the two branches which improves the performance compared to each individual branch.

\textbf{Explainability.}
An inherent property of compositional models is that it can explain the prediction result, in terms of where it perceives which object parts and where it thinks the object is occluded. We illustrate this property in Figure \ref{fig:exp}. For several test images we illustrate five parts which the compositional model has detected with highest likelihood (left) and shows some examples of image patches from the training images which activate the part model most (center). Using theses visualizations, the compositional model can provide an intuitive explanation of why it perceives a certain object in the input image.

	\vspace{-0.1cm}
\section{Conclusion}
\vspace{-0.2cm}
Our extensive experimental results demonstrate that DCNNs cannot recognize partially occluded objects well, if they have not been exposed to partial occlusion during training. 
Even if they have been exposed to severe occlusion during training, they do not generalize well when the spatial distribution or the appearance of the occluders was biased.
In order to resolve these fundamental limitations, we have proposed to combine compositional models and DCNNs. 
In this context, we made the following contributions:

\textbf{Learning of compositional models from DCNN features.} Previous work focused on learning compositional models from plain image pixels, which requires modeling of complex physical processes such as e.g. local deformation or illumination. DCNN features are robust to such  nuisances. Hence, learning compositional models form DCNN features enables us to represent complex objects in natural scenes, which is difficult to achieve with related approaches.

\textbf{Generalizing compositional models to 3D objects.} We propose to use mixtures of compositional models for representing 3D objects. Our experimental results show that mixtures outperform single compositional models at object classification.

\textbf{Combining compositional models and deep networks.} We combine compositional models and DCNNs and demonstrate that they outperform a standard deep network at object classification under partial occlusion by $19.4\%$ on MNIST digits and $13.9\%$ on objects from the PASCAL3D+ dataset in \textit{absolute} classification performance.

\textbf{Acknowledgement.} This work was supported by the Swiss National Science Foundation with grant P2BSP2 181713 and the Office of Naval Research with grant N00014-18-1-2119.
	
	{\small
		\bibliographystyle{plain}
		\bibliography{egbib}
	}

\end{document}